\pdfoutput=1

\documentclass[11pt]{article}
\usepackage{float}
\usepackage{acl2021}
\usepackage{pifont}

\usepackage{times}
\usepackage{latexsym}
\usepackage{subfigure}

\usepackage[T1]{fontenc}

\usepackage[utf8]{inputenc}

\usepackage{microtype}
\usepackage{amsmath, amssymb}
\usepackage{booktabs}
\usepackage{graphicx}
\usepackage{multirow}
\usepackage[noend]{algpseudocode}
\usepackage{amsmath}
\usepackage{algorithmicx,algorithm}
\let\oldtheenumi=\thefootnote

\title{Guiding AMR Parsing with Reverse Graph Linearization}

\author{
 Bofei Gao$^1$$\footnotemark[1]$, Liang Chen$^1$$\footnotemark[1]$, \ Peiyi Wang$^1$, \textbf{Zhifang Sui$^1$,\  Baobao Chang$^1$$\footnotemark[2]$} \\
\textsuperscript{1} National Key Laboratory for Multimedia Information Processing, 
\\ School of Computer Science, Peking University \\
 \leftline{\texttt{\qquad\qquad gaobofei@stu.pku.edu.cn\qquad leo.liang.chen@outlook.com}} \\
 \leftline{\texttt{\qquad\qquad wangpeiyi9979@gmail.com\qquad\quad\{szf,chbb\}@pku.edu.cn}}
}

\date{}

\begin{document}

\newcommand{\cl}[1]{\textcolor{blue}{#1 \\--cl}}
\newcommand{\peiyi}[1]{\textcolor{orange}{\bf \small [ #1 --Peiyi]}}

\maketitle
\begin{abstract}


Abstract Meaning Representation (AMR) parsing aims to extract an abstract semantic graph from a given sentence. The sequence-to-sequence approaches, which linearize the semantic graph into a sequence of nodes and edges and generate the linearized graph directly, have achieved good performance. However, we observed that these approaches suffer from structure loss accumulation during the decoding process, leading to a much lower F1-score for nodes and edges decoded later compared to those decoded earlier. To address this issue, we propose a novel Reverse Graph Linearization (RGL) enhanced framework. RGL defines both default and reverse linearization orders of an AMR graph, where most structures at the back part of the default order appear at the front part of the reversed order and vice versa. RGL incorporates the reversed linearization to the original AMR parser through a two-pass self-distillation mechanism, which guides the model when generating the default linearizations. Our analysis shows that our proposed method significantly mitigates the problem of structure loss accumulation, outperforming the previously best AMR parsing model by 0.8 and 0.5 Smatch scores on the AMR 2.0 and AMR 3.0 dataset, respectively. The code are available at \url{https://github.com/pkunlp-icler/AMR_reverse_graph_linearization}.
\end{abstract}
\renewcommand{\thefootnote}{\fnsymbol{footnote}}
\footnotetext[1]{Equal Contribution.}
\footnotetext[2]{Corresponding Author.}

\renewcommand{\thefootnote}{\oldtheenumi}
\section{Introduction}

\begin{figure}[!t]
    \centering
    \includegraphics[width=1\linewidth]{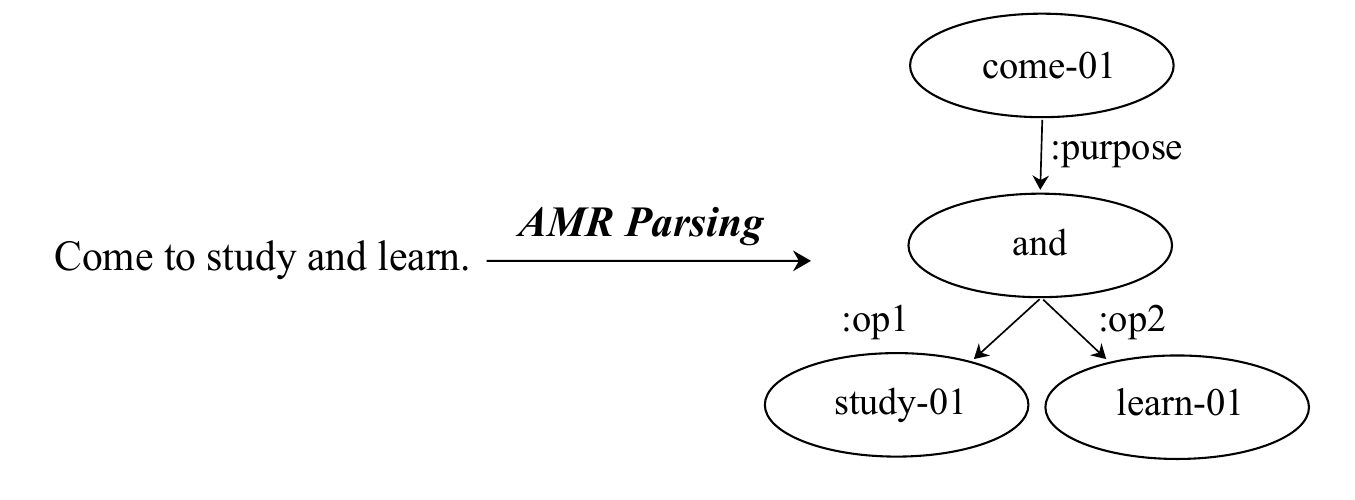}
    \caption{An example of AMR Parsing of the sentence ``Come to study and learn''.}
    \label{fig:amr_example}
\end{figure}

Abstract Meaning Representation (AMR) \citep{ban-AMR} is a formalization of a sentence's meaning using a directed acyclic graph that abstracts away from shallow syntactic features and captures the core semantics of the sentence. AMR parsing involves transforming a textual input into its AMR graph, as illustrated in Figure~\ref{fig:amr_example}. Recently, sequence-to-sequence (seq2seq) based AMR parsers~\citep{xu-seqpretrain,bevil-spring,HCL,bai-etal-2022-graph,Seq2SeqAP,chen-etal-2022-atp, BIBL} have significantly improved the performance of AMR parsing. In these models, the AMR graph is first linearized into a token sequence during traditional seq2seq training, and the output sequence is then restored to the graph structure after decoding. AMR parsing has proven beneficial for many NLP tasks, such as summarization~\citep{liao-amr-tm, hardy-amr-tm}, question answering~\citep{mitra-amr-qa, sacha-amr-qa}, dialogue systems~\citep{Bonial2020DialogueAMRAM,Bai2021SemanticRF}, and information extraction~\citep{rao-amr-ie, wang-amr-ie, zhang-amr-ie,Xu2022ATA}.

\begin{figure}[!t]
\centering
\subfigure[Node prediction]{
\includegraphics[width=0.4\linewidth]{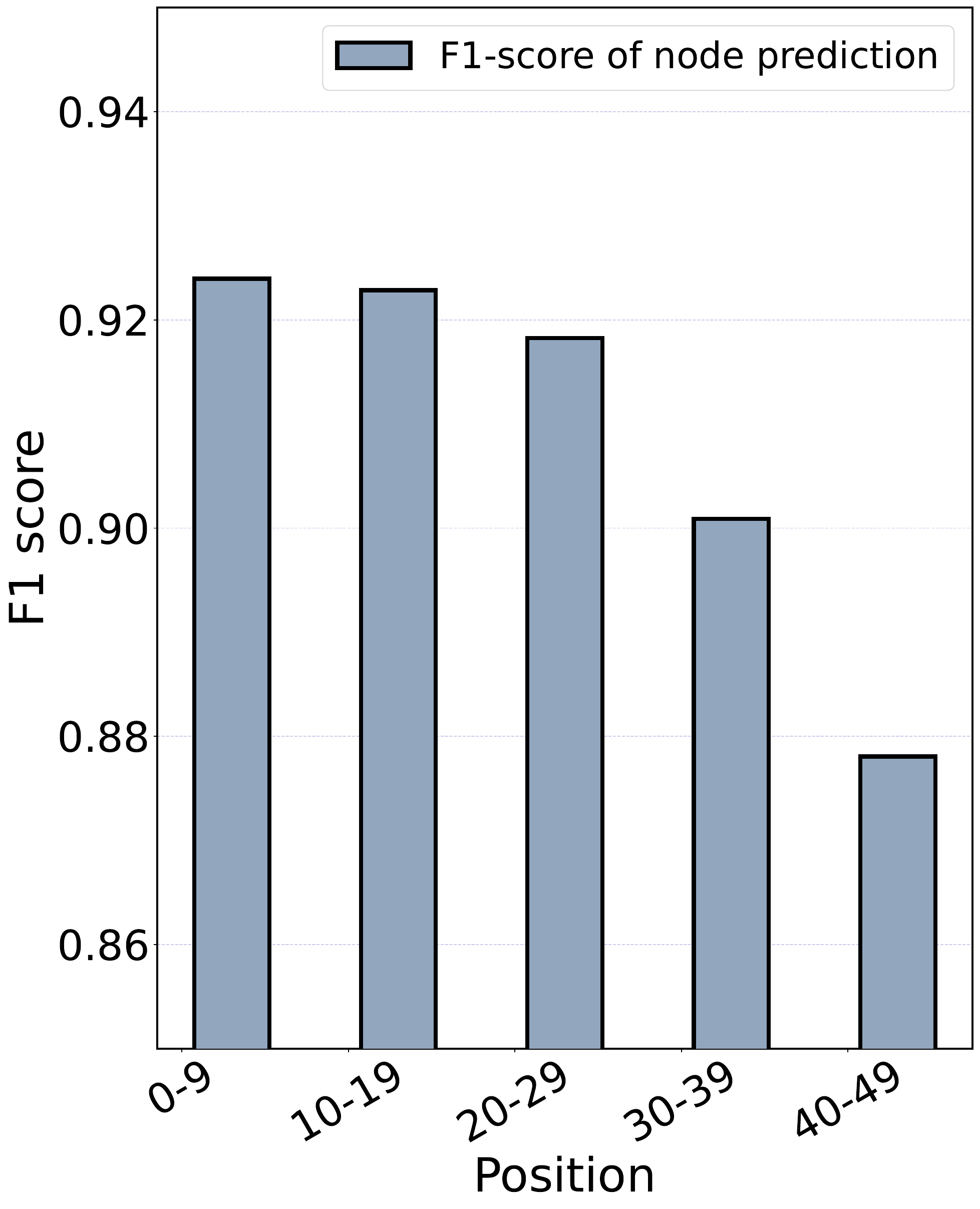}
}\subfigure[Relation prediction]{
\includegraphics[width=0.4\linewidth]{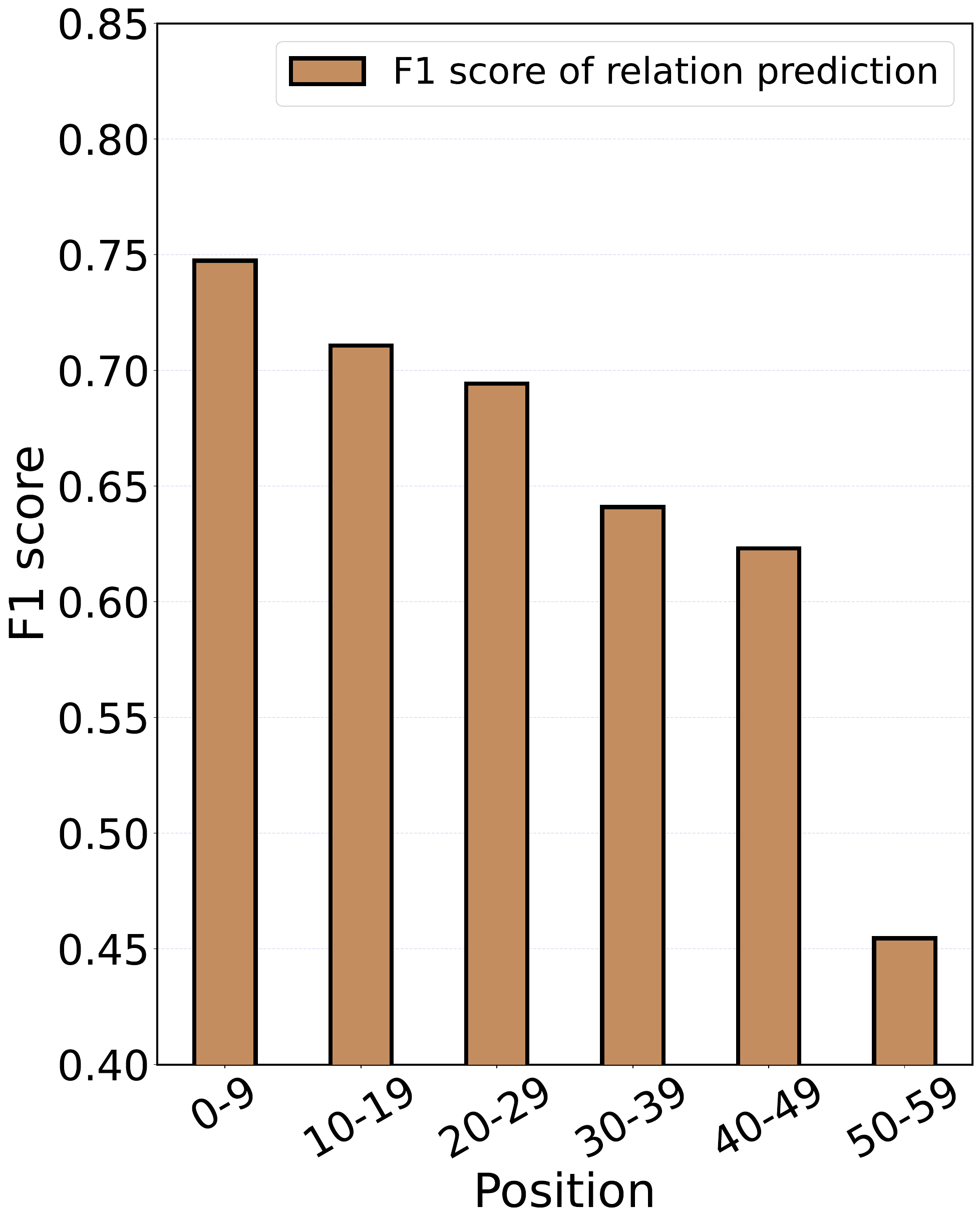}
}
\caption{There is a negative correlation  between the F1-score of the node or relation prediction and the position. The results are obtained from AMRBART \cite{bai-etal-2022-graph} on the test set of AMR 2.0.}
\label{fig:graph struture loss}
\end{figure}

In this study, we aim to address the issue of structure loss accumulation in seq2seq-based AMR parsing. Our analysis (Figure~\ref{fig:graph struture loss}) shows that the F1-score of structure prediction (node and relation) decreases as the generation direction progresses.  This phenomenon is a consequence of the error accumulation in the auto-regressive decoding process, a common problem in natural language generation~\citep{ing2007accumulated, zhang-etal-2019-bridging,LiuScheduled2021}. 

However, unlike natural language, the linearization of AMR graphs does not follow a strict order, as long as the sequence preserves all nodes and relations in the AMR graph. To this end,  we define two linearization orders based on the depth-first search (DFS) traversal, namely Left-to-Right (L2R) and Right-to-Left (R2L). The L2R order is the conventional linearization used in most previous works~\citep{bevil-spring,bai-etal-2022-graph,chen-etal-2022-atp}, where the leftmost child corresponding to the penman annotation is traversed first.  In contrast, the R2L order is its reverse, where the structures at the end of the L2R order appear at the beginning of the R2L order. By training AMR parsing models with R2L linearization, it improves the accuracy of predictions for the structures at the end of the L2R order, which are less affected by the accumulation of structure loss.

We propose to enhance AMR parsing with reverse graph linearization (RGL). Specifically, we incorporate an additional encoder to integrate the reverse linearization graph and replace the original transformer decoder with a mixed decoder that utilizes gated dual cross-attention, taking input from both the hidden states of the sentence encoder and the graph encoder. We design a two-pass self-distillation mechanism to prevent the model from overfitting to the gold reverse linearized graph as well as to further utilize it to guide the model training. Our analysis shows that our proposed method significantly mitigates the problem of structure loss accumulation, outperforming the previously best AMR parsing model \citep{bai-etal-2022-graph} by 0.8 Smatch score on the AMR 2.0 dataset and 0.5 Smatch score on the AMR 3.0 dataset.

Our contributions can be listed as follows:

 1. We explore the structure loss accumulation problem in sequence-to-sequence AMR parsing.

2. We propose a novel RGL framework to alleviate the structure loss accumulation by incorporating reverse graph linearization into the model, which outperforms previously best AMR parser.

3. Extensive experiments and analysis demonstrate the effectiveness and superiority of our proposed method.





\section{Backgrounds}
\subsection{Seq2Seq based AMR Parsing}
In our work, we followed previous methods \cite{ge-seq2seqamr, bevil-spring, bai-etal-2022-graph}, which formulate AMR parsing as a sequence-to-sequence generation problem.
Formally, given a sentence $\mathbf{x} = (x_1, x_2, ..., x_N)$, the model needs to generate a linearized AMR graph $\mathbf{y}=(y_1, y_2, ..., y_M)$ in an auto-regressive manner.

Assuming that we have a training set containing $N$ sentence-linearized graph pairs $(x_i, y_i)$, the total training loss of the model is computed by the cross-entropy loss which is listed as follows:

\begin{equation}\label{eq_ce}
    L_{CE} = -\sum_{i=1}^N\sum_{t=1}^{m_i}log p(y_t^i|y_{<t}^i, x^i)
\end{equation}

where $m_i$ is the length of $i^{th}$ linearized AMR graph, and $y_{<t}^i$ is the previous tokens.


\subsection{Graph Linearization Order}
\label{subsec:R2L_introduction}
\begin{table}[t]
    \centering
    \resizebox{0.45\textwidth}{!}{
\begin{tabular}{lc}
        \toprule
              Direction & Linearized AMR Graph \\
        \midrule
         \textbf{Left-to-Right} & (c/come-01 :purpose (a/and :op1 (s/study-01) :op2 (l/learn-01))) \\
        \midrule
         \textbf{Right-to-Left} & (c/come-01 :purpose (a/and :op2 (l/learn-01) :op1 (s/study-01)))\\
        \bottomrule
    \end{tabular}}
    \caption{The AMR graph shown in Figure \ref{fig:amr_example} with different linearization order. "Left-to-Right" follows the standard DFS traversal order. "Right-to-Left" follows the reverse DFS traversal order.}
    \label{tab:gf}
\end{table}

As shown in Table~\ref{tab:gf}, we formalize two types of graph linearization, the corresponding AMR graph is shown in Figure \ref{fig:amr_example}. \textbf{Left-to-Right (L2R)} denotes that when we use the depth-first search (DFS) to traverse the children of a node, we first start from the leftmost child and then traverse to the right, which is identical to the order of penman annotation and is the default order of sequence-to-sequence based AMR parsers \cite{bevil-spring, bai-etal-2022-graph, chen-etal-2022-atp}. In contrast, \textbf{Right-to-Left (R2L)} traverses from the rightmost child to the leftmost child, which is the reverse of the standard traversal order. When the input sentence is long or contains multi-sentence, most of the nodes or relationships that are positioned later in the L2R sequence will appear earlier in the R2L sequence.

\begin{figure*}[!ht]
    \centering
    \includegraphics[width=1\linewidth]{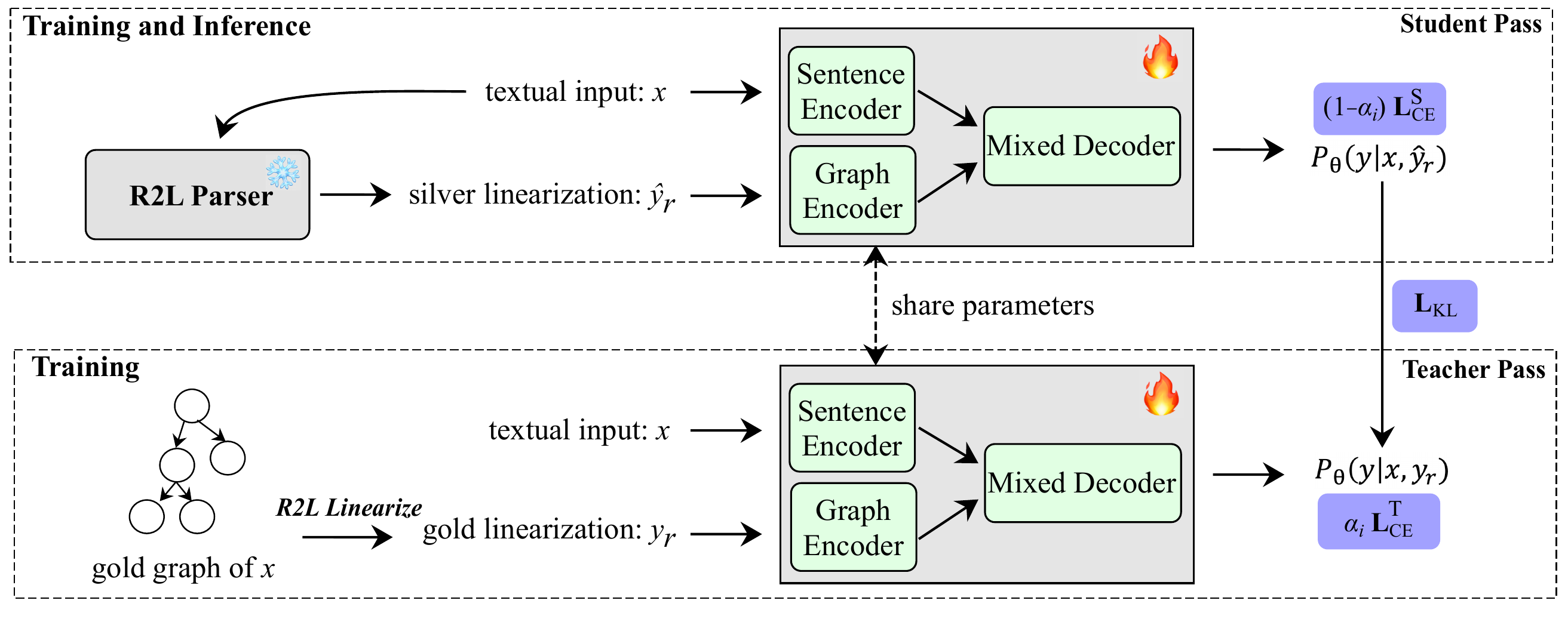}
    \caption{The overview of our method. In addition to the encoder-decoder model, an additional graph encoder is used to incorporate reverse graph linearization. Following the paradigm of self-distillation, we regard the model with the input of the gold linearization $y_r$ and $x$ as the teacher model and $\hat{y}_r$ parsed by a pre-trained R2L parser and $x$ as the student model. The model does twice forward pass to obtain the output probabilities of the teacher and the student in each training step. We calculate the cross-entropy loss of teacher and student as well as their KL divergence as the training loss. Given a sentence $x$ during inference, the model generates the standard AMR linearization using $x$ and its silver linearization $\hat{y}_r$.}
    \label{fig:RGO}
\end{figure*}

\section{Methodology}
\label{sec:method}


\subsection{Overview}
 Our method is illustrated in Figure \ref{fig:RGO}. In addition to the traditional encoder-decoder architecture, we have incorporated a graph encoder to include the reverse linearization sequence. As a result, the model now takes both the sentence and its reverse linearization as input. We modify the original transformer decoder with a mixed decoder that uses gated dual cross-attention in each decoder layer, allowing the integration of hidden representations from both the sentence encoder and the graph encoder. During inference, we need an additional R2L AMR parser that generates the reverse linearization $\hat{y}_r$ of the sentence and then feed both the input sentence $x$ and $\hat{y}_r$ to the model.
 
To obtain reverse linearization during training, a common intuitive approach is to linearize the gold AMR graph into the gold reverse linearization, denoted by $y_r$. However, simply using $y_r$ and the source sentence $x$ as input for all training data can lead to overfitting of the model to $y_r$, causing it to ignore the importance of the source sentence. As a result, the model may simply copy from $y_r$ and generate $y$ during training. This can limit the model's performance during inference due to the noise introduced by the generated reverse linearization, denoted by $\hat{y}_r$.
 

 To prevent the model from overfitting to $y_r$ we introduce silver linearization $\hat{y}_r$ during training. While we still hope to utilize the gold linearization $y_r$  to guide the training, we design a two-pass self-distillation mechanism. Alongside $y_r$, we incorporate $\hat{y}_r$, which is parsed by the additional R2L AMR parser during training. The teacher model takes $y_r$ and $x$ as input, while the student model takes $\hat{y}_r$ and $x$. During each training step, the model performs two forward passes and computes cross-entropy losses, $L_{CE}^T$ for the teacher and $L_{CE}^S$ for the student. We employ KL divergence $L_{KL}$ to guide the student with the teacher's output. We also design a loss scheduler to balance the weight $\alpha_i$  for $L_{CE}^T$ and $L_{CE}^S$ at optimization step $i$. 

\subsection{Model Structure}
\label{sec: model overview}

As shown in Figure \ref{fig:RGO}, our model mainly consists of three parts: sentence encoder, graph encoder, and mixed decoder. The major structural difference from standard pretrained models, e.g. BART \cite{lew-bart}, is that we use a graph encoder to integrate the reverse linearized structural information to guide the model. 


\paragraph{Sentence Encoder}
The sentence encoder receives the given sentence $s = (s_1, s_2, ..., s_N)$, and encodes it to the hidden representations $H^s=(\textbf{h}_1^s, \textbf{h}_2^s, ..., \textbf{h}_N^s)$, which is the same as the encoder of pretrained transformer models.

\paragraph{Graph Encoder}
Following \cite{bevil-spring, bai-etal-2022-graph}, we adopt the standard transformer encoder to encode the structural information. Given the reverse-linearized AMR graph, the output of the graph encoder is $H^g=(\textbf{h}_1^g, \textbf{h}_2^g, ..., \textbf{h}_M^g)$. 

\begin{figure}[t]
    \centering
    \includegraphics[width=0.9\linewidth]{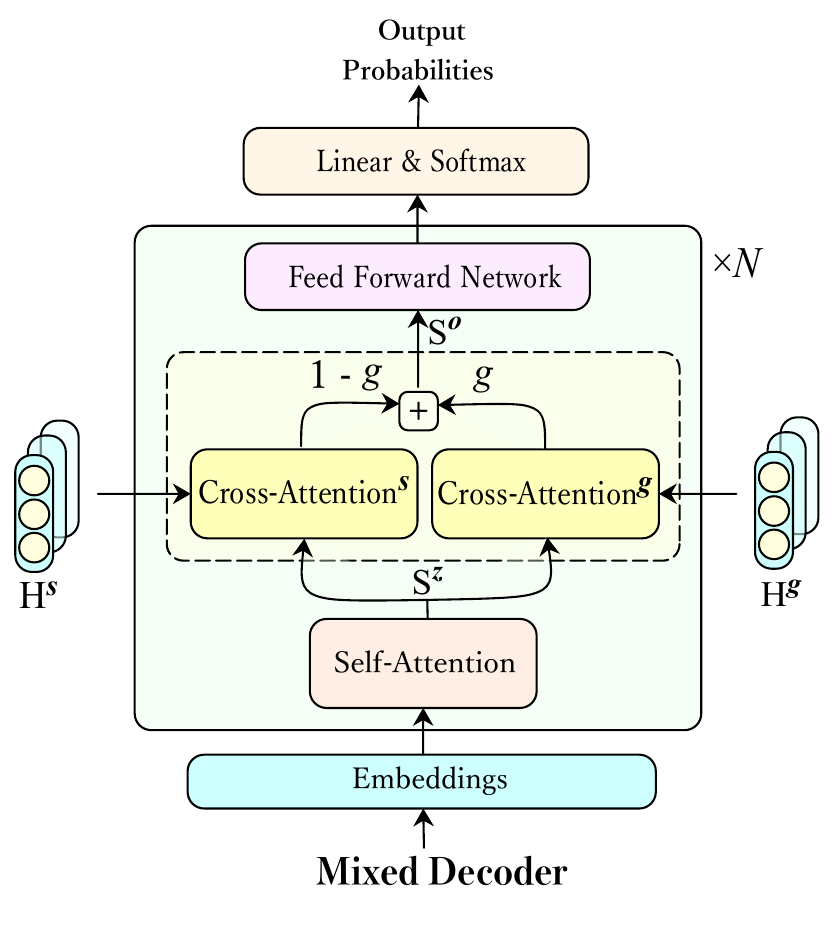}
    \caption{The illustration of the mixed decoder in RGL. $H^s$ and $H^g$ are the hidden representations from the sentence encoder and graph encoder. 
     The module enclosed by the dashed line is the gated dual cross-attention, which integrates the outputs of the dual attention through a gate predicted by an FFN. For brevity and focus, the residual connection and normalization are omitted from the figure.}
    \label{fig:mixed_decoder}
\end{figure}

\paragraph{Mixed Decoder}
Different from the traditional decoder, the mixed decoder takes the hidden states of the sentence $H^s$ and the graph $H^g$ via a gated dual cross-attention layer as shown in Figure \ref{fig:mixed_decoder}. The gated dual cross-attention layer contains two cross-attention modules which are used to integrate $H^s$ and $H^g$ respectively. In the decoder layer, the output of the self-attention module is $S^z \in \mathbb{R}^{k \times d}$, where $k$ is the number of tokens in the decoder input and $d$ is the size of the hidden state. The output of each cross-attention module can be computed as:
\begin{equation}\label{eq_mth_s}
    S^s = \text{CrossAttn}^s(S^z, H^s, H^s)
\end{equation}
\begin{equation}\label{eq_mth_g}
    S^g = \text{CrossAttn}^g(S^z, H^g, H^g)
\end{equation}
where the two cross-attention modules contains the same query $S^z$ but different key-value $H^s$ and $H^g$ respectively.

The output of the gated dual cross-attention module $S^o$ is the weighted sum of $S^s$ and $S^g$.
$$
    S^{o} = \bold{g} \cdot S^g + (\bold{1-g}) \cdot S^s
$$
where $\bold{g} \in \mathbb{R}^{K \times 1}$ is predicted by a feed-forward network:
\begin{equation}\label{eq_mth_gi}
    \bold{g} = \sigma(\mathbf{V}^\mathrm{T} \text{tanh}(\mathbf{W}^\mathrm{T}S^z + b_1) + b_2)
\end{equation}
$\sigma$ is the sigmoid function, $\mathbf{W}\in\mathbb{R}^{d\times d}$ , $\mathbf{V}\in\mathbb{R}^{d\times 1}$, $b_1 \in \mathbb{R}^{d\times 1}$ and $b_2 \in \mathbb{R}$ are trainable parameters and bias.



\subsection{Training Objective}
\label{sec: training}

The training objective of the RGL is:
\begin{equation}
    L = \alpha_i L_{CE}^T + (1 - \alpha_i) L_{CE}^S + L_{KL}
\end{equation}
where $\alpha_i$ is a balancing weight related to $i_{th}$ iteration. $L_{CE}^T$ and $L_{CE}^S$ are the cross-entropy loss of the teacher and the student respectively and $L_{KL}$ is the self-distillation loss.


\paragraph{Self-distillation}
To further guide the model with gold reverse linearization $y_r$ during training as well as to avoid the model from overfitting to it and ignoring the sentence $x$, we propose a two-pass self-distillation mechanism during training. As shown in Equation \ref{eq_teacher} and \ref{eq_student}, we regard the forward pass taking $y_r$ as input a teacher and $\hat{y}_r$ as a student. To obtain the output distribution of both the teacher and the student, the model performs two forward passes in one training step. Note that the teacher and the student model share the same parameters. 
\begin{equation}\label{eq_teacher}
    \resizebox{.87\hsize}{!}{$p(y|x, y_r) = \prod\limits_{i=1}^Mp(y_i|(y_1,y_2,...,y_{i-1}),x, y_r)$}
\end{equation}
\begin{equation}\label{eq_student}
    \resizebox{.87\hsize}{!}{$q(y|x, \hat{y_r}) = \prod\limits_{i=1}^Mp(y_i|(y_1,y_2,...,y_{i-1}),x, \hat{y_r})$}
\end{equation}

To distill the knowledge from the teacher pass to the student pass, we guide the output of student pass with the teachers by minimizing the Kullback–Leibler divergence loss:

\begin{equation}\label{eq_self_distill}
    L_{KL}(p,q) = \sum_{i=1}^{D}p_ilog(\frac{p_i}{q_i})
\end{equation}
where $p$ and $q$ are the output probabilities of the teacher and the student respectively, $D$ is the number of classes which is the total size of the target vocabulary.

\paragraph{Loss scheduler}
Inspired by the idea of curriculum learning, we introduce a loss scheduler to better balance the training process. We set an adaptive coefficient $\alpha_i$ to control the weights of $L_{CE}^T$ and $L_{CE}^S$. $\alpha_i$ gradually decays with the increase of training step $i$. The model is supposed to learn more from gold linearization when its capability is weak so that the model can converge quickly. When the model's capability is strong, it is supposed to have the ability to infer from the noisy silver linearization, which can make the model more capable and robust to noise during inference since we do not have a gold linearization graph during inference. The $\alpha_i$ can be computed as exponential decay:
\begin{equation}
    \alpha_i = k_1 * e^{-k_2 *i}, 0 \leq i \leq total\_steps
\end{equation}
where $k_1$ and $k_2$ are hyper-parameters that can control the upper- and lower-bounds of the $\alpha_i$. We set the upper bound of $\alpha_i$ to 0.8 and the lower bound to 0.2 without further tuning.

\subsection{Inference}
\label{sec: inference}
Given a sentence, we first use the R2L AMR parser to generate its reverse linearization. Then the trained RGL model takes the reverse linearization and the sentence as input and decodes the standard L2R AMR linearization.

\begin{table*}[!t]
    \centering
    \resizebox{0.9\textwidth}{!}{%
    \begin{tabular}{llcccccccccc}
        \toprule
                ~& Model & \textsc{SMATCH} & NoWSD & Wiki & Conc. & NER & Neg. & Unll. & Reen. & SRL \\
        \midrule

        \multirow{8}{*}{\rotatebox[origin=c]{90}{{AMR 2.0}}}    
        
                                          
        ~& SPRING \citep{bevil-spring}          & 83.8  & 84.4& 84.3 & 90.2 & 90.6 & 74.4  & 86.1 & 70.8 & 79.6 \\

         ~&SPRING (w/ silver) \citep{bevil-spring}        & 84.3 & 84.8& 83.1 & 90.8 & 90.5 & 73.6  &  86.7 &  72.4 & 80.5 \\
        
        ~& ATP \citep{chen-etal-2022-atp} & 85.2 & 85.6 & 84.2 & 90.7& \textbf{93.1} & 74.9 & 88.3 & 74.7 & \textbf{83.3} \\
        
        ~& AMRBART \citep{bai-etal-2022-graph} & 85.4 & 85.8 & 81.4 & 91.2 & 91.5 & 74.0 & 88.3 & 73.5 & 81.5 \\
        
        ~& AMRBART (ours) & 85.3 & 85.7 & 84.0 & 91.2 & 90.8 & 74.3 & 88.2 & 73.2 & 81.3 \\

        ~& AMRBART+Multitask (ours) & 85.8 & 86.2 & 83.9 & 91.4 & 91.2 & 75.7 & 88.6 & 74.3 & 81.9 \\

        ~& RGL (ours)    & \textbf{86.1}  & \textbf{86.4} & \textbf{84.5} & \textbf{91.5} & 91.7 & \textbf{76.1}  & \textbf{88.9} & \textbf{74.8} & 82.1 \\

        \midrule
        \multirow{6}{*}{\rotatebox[origin=c]{90}{{AMR 3.0}}}
        ~& SPRING (w/ silver) \citep{bevil-spring} & 83.0 & 83.5& 82.7 & 89.8 & 87.2 & 73.0  & 85.4 & 70.4 & 78.9 \\
        ~& ATP \citep{chen-etal-2022-atp} & 83.9 & 84.3 & 81.0 &89.7 &88.4 & \textbf{73.9}  & 87.0 & \textbf{73.9} & \textbf{82.5} \\
        ~& AMRBART \citep{bai-etal-2022-graph} & 84.2 & 84.6 & 78.9 & 90.2 & \textbf{88.5} & 72.1 & 87.1 & 72.4 & 80.3 \\

        ~& AMRBART (ours) & 84.2 & 84.6 & \textbf{83.3} & 90.1 & 88.2 & 73.2 & 87.1 & 71.9 & 80.0 \\

        ~& AMRBART+Multitask (ours) & 84.4 &84.7&82.9&90.3&88.1&73.1&87.3&72.9&80.4 \\
        
        ~& RGL (ours) & \textbf{84.7} & \textbf{85.1} & 82.8 & \textbf{90.5} & 88.2 & 72.3 & \textbf{87.5} & 73.2 & 80.8 \\

        \bottomrule
    \end{tabular}
    } 
    \caption{\textsc{smatch} and fine-grained F1 scores on AMR 2.0 and 3.0. RGL outperforms AMRBART(ours) significantly with $p < 0.001$ for both AMR 2.0 and AMR 3.0.}
    \label{tab:main_results2.0}
\end{table*}

\section{Experiments}

\subsection{Datasets}
\label{sec: datasets}
We conducted our experiments on two AMR benchmark datasets, AMR 2.0 and AMR 3.0. AMR 2.0 contains $36521$, $1368$, and $1371$ sentence-AMR pairs in training, validation, and testing sets, respectively. AMR 3.0 has $55635$, $1722$, and $1898$ sentence-AMR pairs for training validation and testing set, respectively.

\subsection{Evaluation Metrics}
\label{sec: eval_metrics}
We use the Smatch \citep{cai-smatch}  and further the fine-grained scores \cite{dam-smatch-incremental} to evaluate the performance. The detailed explanations of the metrics are shown in Appendix \ref{sec:evaluatio_metrics}.

 BLINK \citep{wu2019scalable} is used to add wiki tags to the predicted AMR graphs in all the systems in our experiments. We do not apply any re-category methods and other post-processing methods which are the same with \citet{bai-etal-2022-graph} to restore AMR from the token sequence.

\subsection{Main Compared Systems}
\label{sec: models}
\paragraph{AMRBART} We use the current state-of-the-art sequence-to-sequence AMR Parser proposed by \citet{bai-etal-2022-graph} as our main baseline model.

\paragraph{RGL} We initialize our model using AMRBART \citep{bai-etal-2022-graph}. The sentence encoder and the graph encoder are initialized the same as the AMRBART encoder, but they have individual gradients during training. Full details of the compared systems are listed in Appendix~\ref{sec: training_details}.


\subsection{Main Results}
We report the results of our method with several Seq2seq baselines on two major datasets, AMR 2.0 and AMR 3.0 in table \ref{tab:main_results2.0}. Our method outperforms previous methods significantly and provides a state-of-the-art AMR parser. 

In comparison with the baseline AMRBART, our method outperforms it by \textbf{0.8} Smatch point on AMR 2.0 and \textbf{0.5} Smatch point on AMR 3.0. Moreover, our method does not introduce any additional data and is compatible with existing methods such as \citet{chen-etal-2022-atp} and \citet{bai-etal-2022-graph}.




\subsection{Ablation Study}
\label{subsec:ablation}
\begin{table}[t]
    \centering
    \resizebox{0.3\textwidth}{!}{
\begin{tabular}{lcc}
        \toprule
             Model & \textsc{Smatch} \\
        \midrule
             AMRBART (ours) & 85.3 \\
             RGL (ours) & \textbf{86.1} \\
            \quad - w/o silver linearization & 85.0 \\
            \quad - w/o loss scheduler & 85.9 \\
            \quad - w/o self-distillation & 85.7 \\
        \bottomrule
    \end{tabular}}
    \caption{Ablation study results on the RGL. "w/o loss scheduler": remove the loss scheduler in the training process, where we simply add up all loss terms. "w/o self-distillation": remove the $L_{KL}$ and $L_{CE}^T$ from training objective. "w/o silver linearization": remove the $L_{KL}$ and $L_{CE}^S$ from training objective.}
    \label{tab:ablation_study}
\end{table}

\paragraph{Model Training}
Table \ref{tab:ablation_study} presents the results of an ablation study in which we analyze how different training methods affect the performance of RGL.

We observed a significant drop in model performance when we removed the silver linearization from the training process. This approach involves feeding the model with the gold linearization during training while using the silver linearization at inference. We believe this drop in performance occurred for two reasons. First, since the gold reverse linearization and the target are highly similar in structure, the model can be easily overfitted to the gold reverse linearization and ignore the source sentence. This can cause the model to simply replicate the input $y_r$ to $y$ instead of accurately parsing the sentence to an AMR graph. Second, the lack of a structure loss for the gold AMR sequence during training means that the model does not learn to differentiate the correct part of the graph from the noisy part, which is required during inference. Therefore, without the silver graph during training, our model cannot be effectively trained.

We also observed a significant drop in performance when we removed self-distillation from the training process. This highlights the importance of self-distillation in our method, which helps the model avoid the error information caused by noise in silver graphs during training. Nevertheless, our method still outperformed AMRBART, even without self-distillation, which demonstrates the effectiveness of incorporating the reverse linearization into AMR parsing.

Finally, when we removed the loss scheduler, the performance of the model degraded. This emphasizes the importance of the loss scheduler in balancing the teacher and the student during training and enhancing the performance of our method.

\begin{table}[t]
    \centering
    \resizebox{0.4\textwidth}{!}{
\begin{tabular}{c|ccccc}
        \toprule
             Number of Layers  & 12 & 10 & 8 & 6 & 4 \\
        \midrule
        \textsc{Smatch} & 86.1 & 86.0 & 85.7 & 85.9 & 85.6 \\
        \bottomrule
    \end{tabular}}
    \caption{The influence of different number of layers of graph encoder on AMR 2.0.}
    \label{tab:parameters}
\end{table}

\paragraph{Graph Encoder Size} We conduct an ablation experiment on how does the size of graph encoder influence the parsing performance. As shown in Table \ref{tab:parameters}, we only retain the bottom few layers of the graph encoder and we observe that the performance generally declines when the number of layers decreases. However, even when the graph encoder retains only four layers, our model still outperforms AMRBART, which demonstrates the effectiveness of incorporating reverse graph linearization during training.

\section{Analysis}
\subsection{On the Effect of R2L Linearization}
In this section, we replace the input of the graph encoder with different sequences to validate the effectiveness of R2L linearization, which is shown in the upper parts of Table \ref{tab:relevant}. \ding{172} is the proposed RGL and achieves the best performance of all methods. And we replace the input of the graph encoder with the standard L2R linearization without changing other conditions, which is shown at \ding{173}. Inspired by the ideas of \citet{ zhou2019synchronous, zhou2019sequence_both_size}, which explore decoding from both sides for machine translation, we can directly reverse the entire L2R linearization token sequence as the input of graph encoder instead of the R2L linearization, where all the nodes and relations strictly appear at the opposite of L2R linearization, which is the \ding{174} of Table \ref{tab:relevant}.

Comparing \ding{172} to \ding{173}, we observe a more significant improvement when using R2L linearization. This is because some nodes or relations in R2L linearization are predicted earlier by the R2L parser, resulting in less structure loss and higher accuracy, which serves as a complementary source of information for the model. The result proves the effectiveness of incorporating reverse linearization.

Comparing \ding{172} to \ding{174}, we find that the performance would drop if we replace the R2L linearization with a simple reversed L2R token sequence. We believe the main reason for this is that the dependencies between nodes and relationships within the linearized AMR graphs are highly intricate. Simply reversing the sequence can lead to unexpected changes in the sequence, e.g. referential variables, making it challenging for the model to accurately predict after the inversion. In fact, the parsing performance of the simple reverse parser is only 75.9 Smatch score, which is far less than the baseline model. In contrast, R2L linearization is a more reasonable reverse as it is meaning-equivalent to the original L2R linearization and can reach similar parsing performance to the original L2R parser. 

The combined findings demonstrate that incorporating a reverse order is advantageous for AMR parsing. Moreover, the R2L linearization proves to be a more suitable form compared to reversing the input sequence token by token.

\subsection{On Incorporating R2L Linearization }
\label{subsec:relevant_works}
\begin{table}[t]
    \centering
    \resizebox{0.35\textwidth}{!}{
\begin{tabular}{lcccc}
        \toprule
             Model & \textsc{Smatch} \\
        \midrule
            \ding{172} RGL w/ R2L Linearization & \textbf{86.1} \\
            \ding{173} RGL w/ L2R Linearization & 85.8\\
            \ding{174} RGL w/ reverse sequence & 85.8 \\
        \midrule
             \ding{175} Double-decoder+KL & 85.6 \\
             \ding{176} Multitask & 85.8 \\
             \ding{177} Concatenate Input & 85.3 \\
        \bottomrule
    \end{tabular}}
    \caption{\textsc{smatch} of different reverse linearizations and different integration methods. The upper part compares different graph encoder inputs of the RGL. The lower part compares different ways to incorporate R2L linearization.}
    \label{tab:relevant}
\end{table}

In this section, we compare different methods to incorporate the R2L linearization, including several works in other fields adapted into the setting of AMR parsing, which are shown in the lower part of Table \ref{tab:relevant}.

\paragraph{Double-decoder+KL} \citet{xie2021tree_mutual_learning} using two decoders to generate two different linearizations i.e. DFS and BFS for code generation and leverages the mutual information to narrow the KL-divergence between the outputs. We adapt this method into AMR parsing settings, where the two different linearizations are L2R and R2L. Then we narrow the output distributions of corresponding nodes and relations of the two linearizations.

\paragraph{Multitask} A simple method to integrate extra linearization order is through multitask learning, where the model learns to predict both the L2R and R2L AMR graph. 
During training, a task identifier <L2R> or <R2L> is added to the beginning of the input sentence to differentiate the output's order. 
During inference, we individually test the two orders and select the order with the higher Smatch score (L2R) as the final result. 
The difference from \ding{172} in model architecture is that we share the decoder which learns to generate different linearizations simultaneously, rather than use an extra decoder.

\paragraph{Concatenate Input} Another intuitive way to directly introduce reverse linearization into the model is to concatenate it with the textual input. Compared with RGL, this method reduces the additional graph encoder without changing other conditions.

Experimental results show that both \ding{175} and \ding{176} can benefit the model, which implicitly incorporates the R2L linearization to the model through the training loss. However, the proposed RGL explicitly integrates reverse linearization into the model as the extra input, achieving more significant improvements.

However, integrating the R2L linearization through directly concatenating them as the model input is not as effective as the RGL. One possible reason for this is that the linearized graph and the sentence are different structures and simply concatenating them from the input text and letting the model learn the extra structural information provided by R2L linearization through one encoder is challenging. Therefore, the extra graph encoder is necessary for encoding the R2L linearization.

Overall, this section demonstrates that RGL is an effective method for incorporating reverse linearization into the model.

\subsection{Effect of RGL on structure loss}


\begin{figure}[!tbp]
\centering
\subfigure[Node prediction]{
\includegraphics[width=0.5\linewidth]{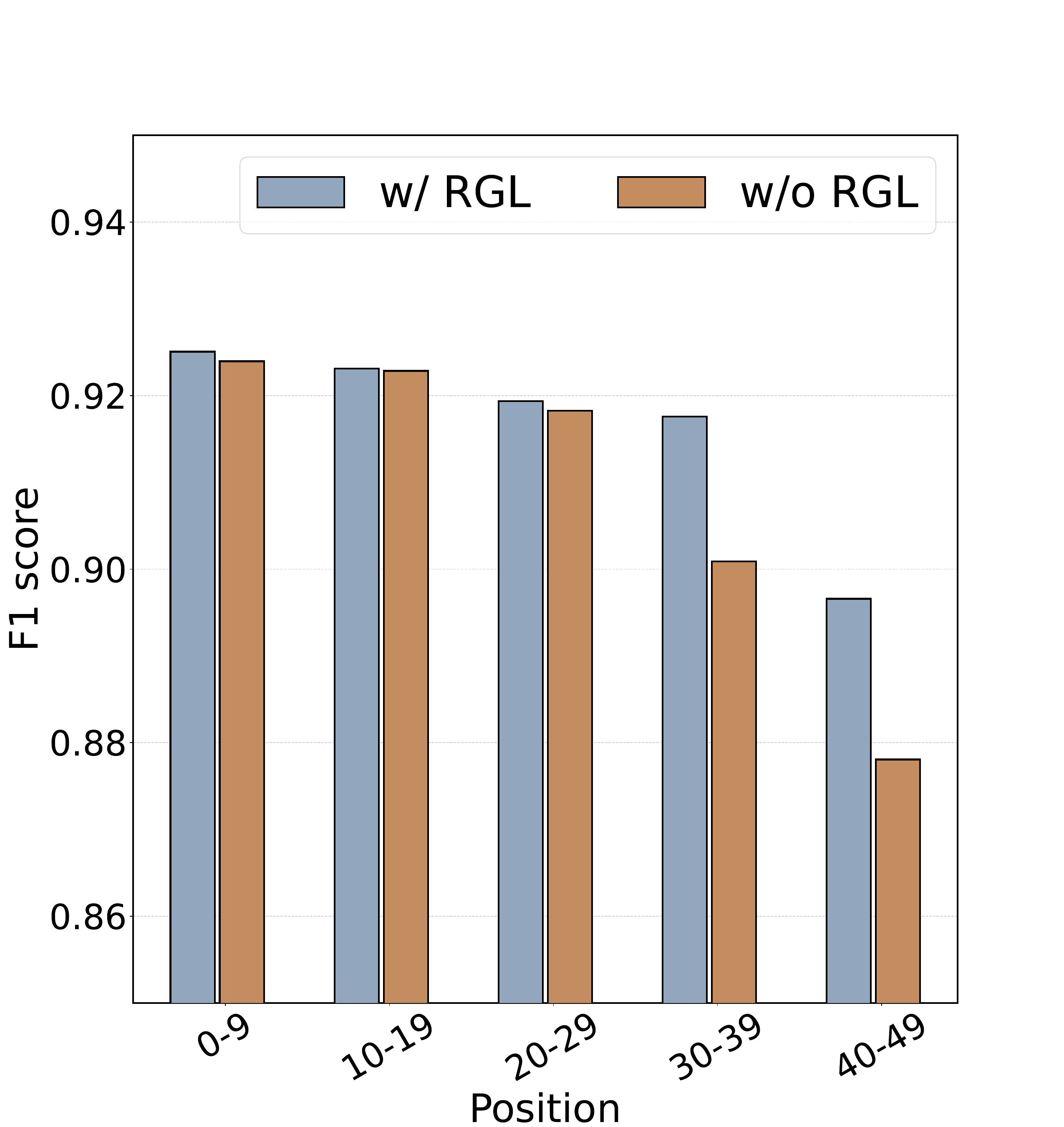}
}\subfigure[Relation prediction]{
\includegraphics[width=0.5\linewidth]{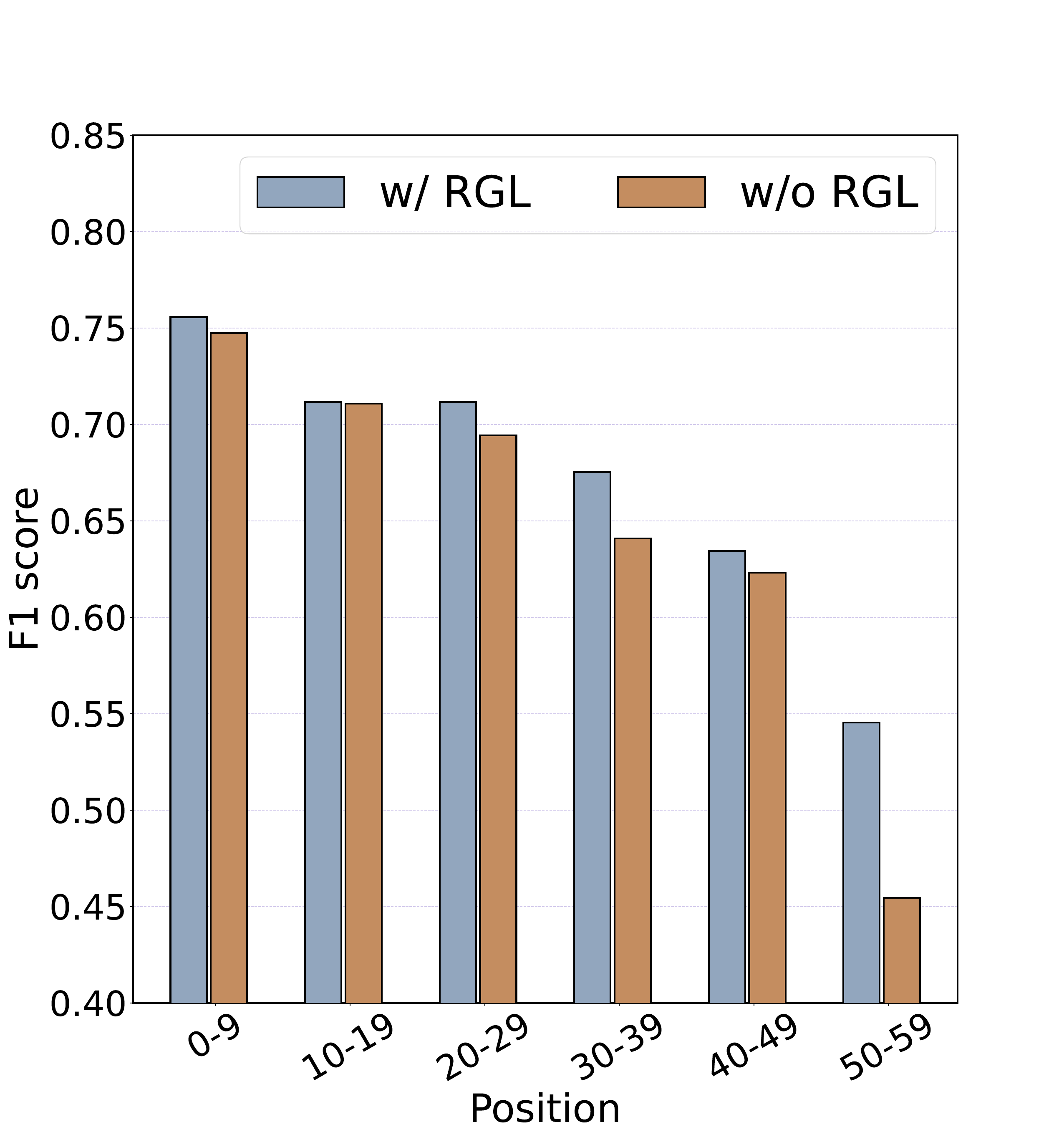}
}
\caption{F1-score of nodes and relations with the increase of the predicted length of AMRBART \citep{bai-etal-2022-graph} represented by orange bars and RGL represented by blue bars.}
\label{fig:effect_RGO}
\end{figure} 


The decrease of F1 scores for nodes and relations with prediction length is shown in Figure \ref{fig:effect_RGO}. 
Compared with the baseline AMRBART, there is a significant improvement in the F1 score of both the node and relation prediction of the RGL when the prediction length is greater than 30. 

To quantify the results, we measured the Pearson coefficients between the F1 scores of nodes and relations and the prediction length. Compared to AMRBART, the Pearson correlation coefficient of node F1 scores with prediction position decreased from -0.42 to -0.26. The coefficient of relation F1 scores with prediction position decreased from -0.72 to -0.6. It proves that the RGL model can indeed alleviate the structure loss problem.

Our analysis also reveals that node prediction is less affected by structure loss accumulation than relation prediction. We believe this is mainly because node prediction in AMR parsing is relatively easier, whereas relation prediction requires correct node predictions as a precondition.

\subsection{Balancing source and reverse linearization}
\begin{figure}[t]
    \centering
    \includegraphics[width=0.7\linewidth]{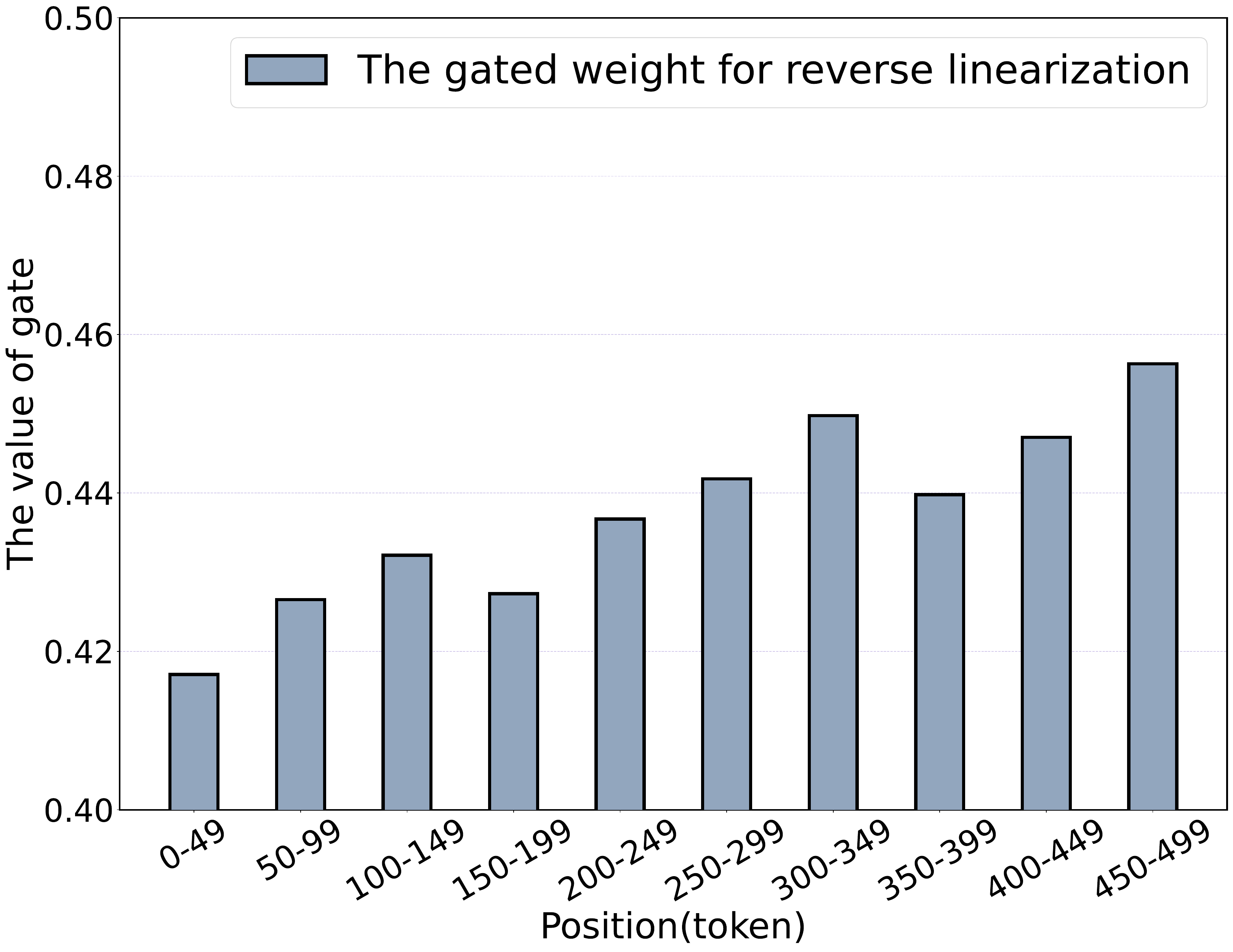}
    \caption{The histogram of the gated weight in the gated dual cross-attention with the increase of the position during inference. A higher value indicates that the model is attending more to the output of the reverse graph encoder in the cross-attention layer. We divided the positions into buckets of size 50 and computed the average gate value across all positions and layers within each bucket, represented by the blue bar in the diagram. }
    \label{fig:gate_with_position}
\end{figure}

Figure \ref{fig:gate_with_position} shows the results of a quantitative analysis of the weight $g$ in the gated dual cross-attention of RGL. We recorded the positions and gated values during model inference on the validation set\footnote{The value range of the x-axis is significantly longer than that of Figure \ref{fig:graph struture loss} because we count all the output tokens in this experiment, instead of picking out tokens representing nodes and relationships from all tokens.} .

The diagram reveals that the average value of the gate is less than 0.5, indicating that the model pays more attention to the source sentence than to the reverse linearization. This suggests that the model is performing sentence-to-AMR conversion, rather than simply copying the reverse linearization.

Furthermore, there is a positive correlation between the gated weight and the position, which provides insight into how our method works. In positions closer to the beginning, the model has greater confidence, resulting in smaller structure loss. The model can predict the AMR graph using only the original source sentence. As the position increases, the model needs to refer to the reverse linearization to compensate for the accumulation of structure loss. Consequently, the gated weight for the reverse linearization becomes larger as the position increases.





\section{Related Work}

AMR parsing aims to convert a textual input to an AMR semantic graph \cite{ban-AMR}. There are mainly four AMR Parsing strategies in previous work, two-stage approaches \cite{flanigan-etal-2014-discriminative,lyu-titov-2018-amr, zhang-etal-2019-amr,zhou-etal-2020-amr}, graph-based approaches \cite{zhang-etal-2019-broad, cai2020amr}, transition-based approaches \cite{naseem2019rewarding, lee2020pushing, fernandez-astudillo-etal-2020-transition, zhou2021amr}, sequence-to-sequence approaches \cite{ge-seq2seqamr, xu-etal-2020-curriculum, bevil-spring, HCL, bai-etal-2022-graph, chen-etal-2022-atp, Seq2SeqAP, BIBL}. In terms of AMR graph linearization, \citet{bevil-spring} explores which linearization method is better for AMR parsing, and \citet{chen-etal-2022-atp} studied how to linearize different semantic resources like SRL to enhance AMR parsing.  Some methods have also been proposed to incorporate graph information into sequence-to-sequence models to compensate for the discrepancy between graph and sequence \cite{yu-gildea-2022-sequence,bai-etal-2022-graph}. While previous seq2seq-based AMR parsing models mostly take the L2R linearization order by default, our work first explores how to leverage different graph linearization orders to enhance AMR parsing.



\section{Conclusion}


In this work, we propose a novel Reverse Graph Linearization (RGL) enhanced framework to address the structure loss accumulation problem observed in the seq2seq-based AMR parsing. Through extensive experiments and analysis, it shows that RGL significantly mitigates the problem of structure loss accumulation and outperforms the previous state-of-the-art model on both AMR 2.0 and AMR 3.0 datasets, which demonstrates the effectiveness of the proposed approach.

\section{Limitation}
Compared to traditional sequence-to-sequence AMR parser, our model needs an additional R2L parser to generate the reverse linearizations, although it can be easily obtained by fine-tuning off-the-shelf AMR parser, e.g. AMRBART \cite{bai-etal-2022-graph} and SPRING \cite{bevil-spring}. Due to the necessity to generate the reverse linearization before AMR parsing, the inference is two times slower than the one-pass AMR parser.

\section{Acknowledgement}
We thank all reviewers for their valuable advice.
This paper is supported by the National Key Research and Development Program of China under Grant No.2020AAA0106700, the National Science Foundation of China under Grant No.61936012 and 61876004.

\section{Ethics Consideration}
We collect our data from public datasets that permit academic use and buy the license for the datasets that are not free. The open-source tools we use for training and evaluation are freely accessible online without copyright conflicts.

\bibliography{acl2021}
\bibliographystyle{acl_natbib}

\clearpage
\appendix

\section{Training Details}
\label{sec: training_details}
\begin{table}[!h]
 \centering
AMR Parsing on AMR 2.0/3.0

 \resizebox{0.4\textwidth}{!}{
  \begin{tabular}{l|c}

  \toprule
  Model Name & AMRBART~\citep{bai-etal-2022-graph} \\
  Pretrained Model & AMRBART-Large \\
  Learning Rate &  8e-6 \\
  Batchsize & 16 \\
  Accumulation Steps & 4 \\
  Max Epochs & 30 \\
  Validation Interval & 1 epoch \\
  Early Stopping & 10 \\
  Beam size & 5 \\
  Warmup Steps & 200 \\
  Entity Linking & BLINK~\citep{wu2019scalable} \\
  
  \bottomrule

  \end{tabular}}

  \caption{The Hyper-Parameters for all of our implemented models including RGL and baseline models.
  }
  \label{tab:hyperp}%
\end{table}

\paragraph{R2L parser} For the R2L parser for inference, we fine-tune AMRBART \citep{bai-etal-2022-graph} using sentences and their corresponding reverse linearized AMR graphs of the training sets.

During training, we also need an R2L parser to parse the sentence into the silver graph of the total training set. If we use the R2L parser exactly the same as that in inference, it will generate silver graphs that are almost the same as the gold graphs, because the R2L parser has already seen all of these data during training. To solve this problem, we use 30$\%$ of the training set (10000 samples in AMR 2.0, 15000 samples in AMR 3.0) to train a ``weaker'' R2L parser, and then use it to infer the entire training set to obtain the silver linearizations\footnote{Gold linearization means that the AMR sequence is obtained by linearization of the gold AMR graph, which is the ground truth AMR graph of the sentence and is free from any errors. Silver linearization means that the AMR graph is parsed from a sentence using an AMR parser, possibly containing noise.} during training. 

We use hyper-parameters shown in table \ref{tab:hyperp} to train all of our implemented models, including the baseline and R2L parser for inference. Before training the RGL, we use the state dict of the encoder of AMRBART to initialize the graph encoder and then train the model using the same configuration. As for the R2L parser for training, we random select a part of the training set in the ratio of 0.3, then we use these gold labeled data to train the R2L parser.

We implemented our models on the Pytorch framework. All the models are trained on a single NVIDIA A100 GPU. Training takes 17 hours on AMR 2.0 and 24 hours on AMR 3.0.

\section{Detailed Evaluation Metrics}
\label{sec:evaluatio_metrics}
We use the Smatch scores \citep{cai-smatch} to evaluate the performance. The further the break down scores \cite{dam-smatch-incremental} is shown as follows.
  i) No WSD, compute while ignoring Propbank senses (e.g., duck-01 vs duck-02), ii) Wikification, F-score on the wikification (:wiki roles), iii) Concepts, F-score on the concept identification task, iv) NER, F-score on the named entity recognition (:name roles), v) Negations, F-score on the negation detection (:polarity roles), vi) Unlabel, compute on the predicted graphs after removing all edge labels, vii) Reentrancy, computed on reentrant edges only, viii) Semantic Role Labeling (SRL), computed on :ARG-i roles only.

\section{Convergence Curve}
\begin{figure}[t]
    \centering
    \includegraphics[width=0.9\linewidth]{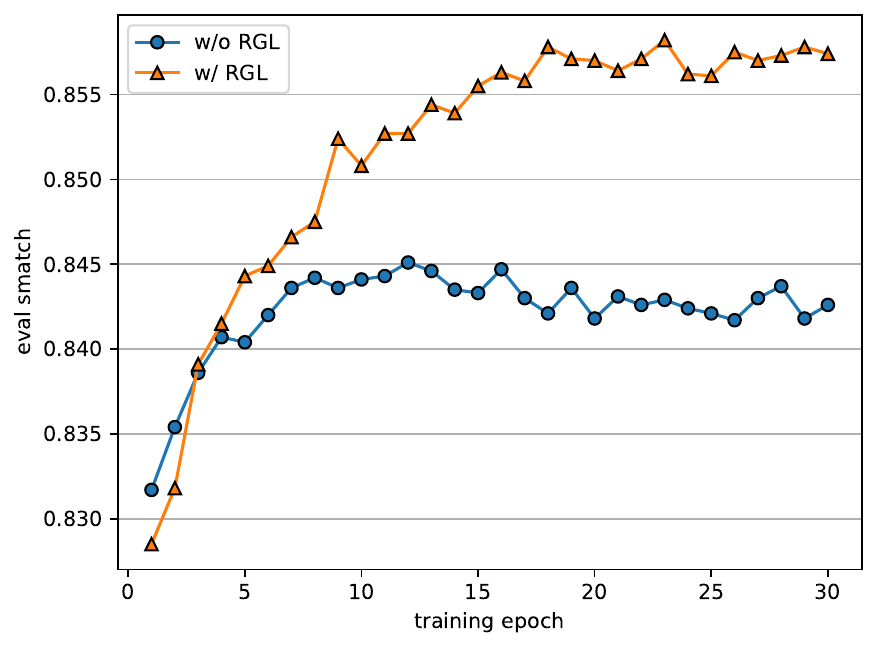}
    \caption{The convergence curve of the RGL and AMRBART.}
    \label{fig:smatch_compare}
\end{figure}

Figure \ref{fig:smatch_compare} presents the convergence curves of RGL and AMRBART on the AMR2.0 dataset. The training process consists of 30 epochs. After each epoch, we compute the SMATCH of RGL and AMRBART on the validation set. Results in Figure \ref{fig:smatch_compare} indicate that RGL outperforms AMRBART significantly.

\section{Error propagation vs. structure loss}
\begin{figure}[!htb]
    \centering
    \includegraphics[width=1\linewidth]{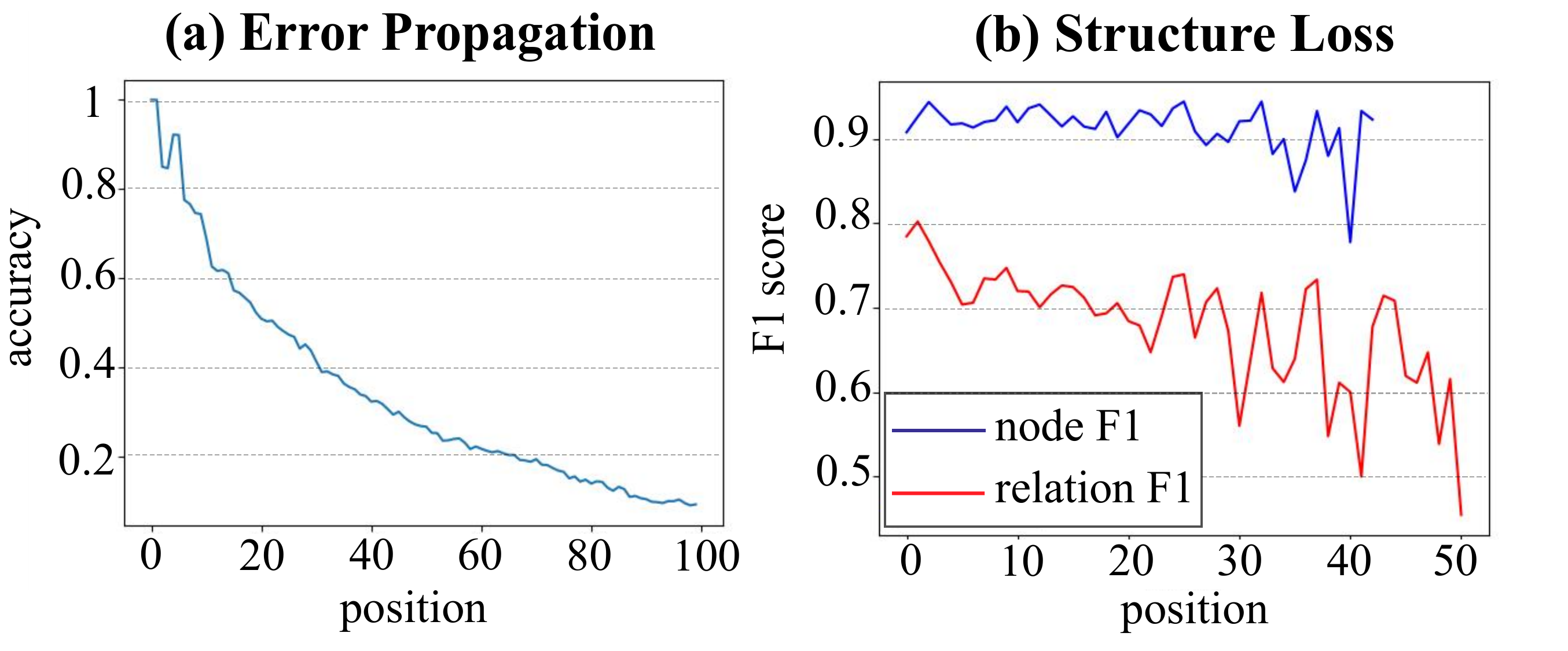}
    \caption{ The descent of (a) position-wise accuracy and (b) graph-wise F1-score of nodes and relations as the decoding progresses. The results are from AMRBART \cite{bai-etal-2022-graph} on the test set of AMR 2.0.  }
    \label{fig:intro_graph_drop}
\end{figure}

Figure \ref{fig:intro_graph_drop} highlights the distinction between error propagation and structure loss. Error propagation is typically evaluated position-wise or within a limited window \cite{LiuScheduled2021}, and is observed in almost every autoregressive method, including sequence-to-sequence based AMR parsing. Once a previous prediction is misplaced or incorrect, subsequent predictions tend to follow the same pattern. In contrast, structure loss evaluates the validity of a node or relation based on its existence in the entire gold graph, rather than its position or window. We argue that structure loss provides a more accurate reflection of the challenges in AMR parsing and other structure generation tasks because it measures the overall quality of the generated AMR graph.

\section{Case Study}
\begin{figure*}[!h]
    \centering
    \includegraphics[width=1\linewidth]{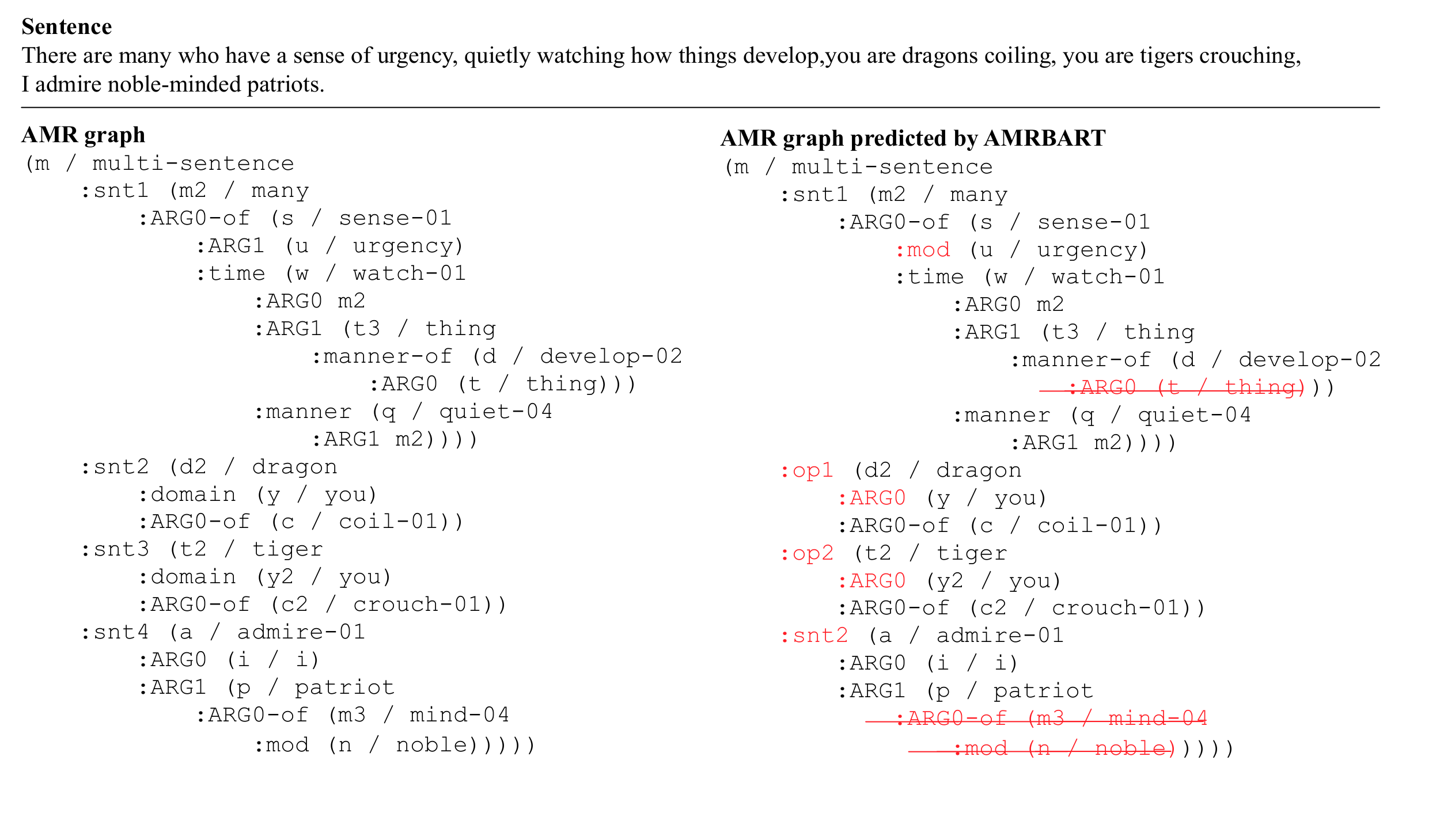}
    \caption{An example of AMR parsing of the long sentence from the validation set of AMR 3.0.}
    \label{fig:case_study}
\end{figure*}

The illustrated example in figure \ref{fig:case_study} shows the accumulation of structural loss more intuitively. We align the variables predicted by the model with the standard AMR graph and mark the prediction errors in red. From the figure, we can see that there are more errors in the later part of the predicted AMR graph. What's more, the relation ":snt2" is wrongly predicted due to the error of the previous relations ":op1" and ":op2", which shows that the duplicate dependencies imposed by sequence-to-sequence manner on AMR parsing have a negative effect.

\end{document}